\documentclass[times, twoside]{zHenriquesLab-StyleBioRxiv}
\usepackage{blindtext}
\usepackage{tabu}
\usepackage{booktabs}

\leadauthor{Hagen} 

\begin{document}

\title{Image Memorability Prediction with Vision Transformers}
\shorttitle{ViTMem}

\author[1,\Letter]{Thomas Hagen}
\author[1,2]{Thomas Espeseth}

\affil[1]{Department of Psychology, University of Oslo, Oslo, Norway}
\affil[2]{Department of Psychology, Oslo New University College, Oslo, Norway}

\maketitle

\begin{abstract}
Behavioral studies have shown that the memorability of images is similar across groups of people, suggesting that memorability is a function of the intrinsic properties of images, and is unrelated to people’s individual experiences and traits. Deep learning networks can be trained on such properties and be used to predict memorability in new data sets. Convolutional neural networks (CNN) have pioneered image memorability prediction, but more recently developed vision transformer (ViT) models may have the potential to yield even better predictions. In this paper, we present the ViTMem, a new memorability model based on ViT, and evaluate memorability predictions obtained by it with state-of-the-art CNN-derived models. Results showed that ViTMem performed equal to or better than state-of-the-art models on all data sets. Additional semantic level analyses revealed that ViTMem is particularly sensitive to the semantic content that drives memorability in images. We conclude that ViTMem provides a new step forward, and propose that ViT-derived models can replace CNNs for computational prediction of image memorability. Researchers, educators, advertisers, visual designers and other interested parties can leverage the model to improve the memorability of their image material.  
\end {abstract}


\section*{Introduction}

Everyone knows that our memories depend on the experiences we have had, facts we have encountered, and the abilities we have to remember them. Combinations of these factors differ between individuals and give rise to unique memories in each of us. However, a complementary perspective on memory focuses on the material that is (to be) remembered rather than the individual that does the remembering. In one central study, Isola et al. \cite{isola2013makes} presented more than 2000 scene images in a continuous repeat-detection task. The participants were asked to respond whenever they saw an identical repeat. The results revealed that the memorability score (percent correct detections) varied considerably between images. Most importantly, by running a consistency analysis in which Spearman’s rank correlation was calculated on the memorability scores from random splits of the participant group, Isola and colleagues \cite{isola2013makes} were able to show that the memorability score ranking was consistent across participants – some images were memorable and some were forgettable. These results indicate that the degree to which an image was correctly detected depended on properties intrinsic to the image itself, not the traits of the observers. This is important because it shows that one can use the memorability scores in a stimulus set to predict memory performance in a new group of participants.

These results have been replicated and extended in a number of studies, revealing that similar findings are obtained with different memory tasks \cite{goetschalckx2018image}, different retention times \cite{goetschalckx2018image, isola2013makes}, different contexts \cite{bylinskii2015intrinsic}, and independent of whether encoding is intentional or incidental \cite{goetschalckx2019incidental}. However, although image memorability has proven to be a robust and reliable phenomenon, it has not been straightforward to pinpoint the image properties that drive it. What seems clear though, is that memorability is multifaceted \cite{bylinskii2022memorability, rust2020understanding}. One way to characterize the underpinnings of memorability is to investigate the contribution from processes at different levels of the visual processing stream. For example, at the earliest stages of processing of a visual scene, visual attributes such as local contrast, orientation, and color are coded. At an intermediate level, contours are integrated, surfaces, shapes, and depth cues are segmented, and foreground and background are distinguished. At a higher level, object recognition is conducted through matching with templates stored in long term memory.

Positive correlations between brightness and high contrast of objects with memorability has been found \cite{dubey2015makes}, but in general, low-level visual factors such as color, contrast, and spatial frequency do not predict memorability well \cite{bainbridge2017memorability, bylinskii2022memorability, isola2011understanding}. This is consistent with results showing that perceptual features are typically not retained in long term visual memory \cite{brady2011review}. In contrast to the low-level features, the evidence for a relation between intermediate to high level semantic features and memorability is much stronger. For example, images that contain people, faces, body parts, animals, and food are often associated with high memorability, whereas the opposite is a typical finding for objects like buildings and furniture and images of landscapes and parks \cite{bylinskii2015intrinsic, dubey2015makes, goetschalckx2019memcat, ICCV15_Khosla}. Other intermediate to high level features such as object interaction with the context or other objects, saliency factors, and image composition also contribute to memorability \cite{bylinskii2022memorability}. Furthermore, although memorability is not reducible to high-level features such as aesthetics \cite{isola2013makes, ICCV15_Khosla}, interestingness \cite{gygli2013interestingness, isola2013makes}, or popularity \cite{ICCV15_Khosla}, emotions, particularly of negative valence, seem to predict higher memorability \cite{isola2011understanding, ICCV15_Khosla}. Finally, memorability seems to be relatively independent of cognitive control, attention, or priming \cite{bainbridge2020resiliency}.

Overall, the available evidence indicates that memorability seems to capture intermediate- to high-level properties of semantics, such as objects or actions, and image composition, such as layout and clutter, rather than low-level features \cite{bylinskii2022memorability, Kramer2022.04.29.490104}. This fits well with the central role of semantic categories in organizing cognition and memory \cite{rosch1976basic}. Generally, the priority of semantic-level information enables us to quickly understand novel scenes and predict future events \cite{medin1998concepts}. For example, when inspecting a novel scene or an image, we do not primarily focus on low-level perceptual features or pixels, but prioritize more abstract visual schemas involving spatial regions, objects, and the relation between them \cite{akagunduz2019defining}. Also, when people are asked to indicate which regions of an image helps them recognize an image, there is high consistency between people’s responses \cite{akagunduz2019defining}. Similarly, fixation map data from eye-tracking have shown that there is a positive correlation between fixation map consistency and scene memorability, and this relation is associated with the presence of meaningful objects \cite{bylinskii2015intrinsic, dubey2015makes, lyu2020overt}. Bylinskii et al. \cite{bylinskii2022memorability} suggest that these properties most efficiently signal information of high utility to our species, for example, emotions, social aspects, animate objects (e.g., faces, gestures, interactions), unexpected events, and tangible objects. 

\subsection*{Memorability prediction}
The finding that the memorability of an image is governed by properties intrinsic to the image itself, not only implies that one can predict memory performance in a new set of participants, as described above, but also that one can predict the memorability of a novel set of images (i.e., memorability is an “image computable” feature). Given the availability of computational algorithms and high-quality training sets of sufficient size, one can predict memorability in novel sets of images for future (or already conducted) behavioral or neuroimaging studies. Such memorability prediction could also be valuable in a number of applied settings (e.g., within education, marketing and human-computer interaction).

Memorability researchers have employed computer vision models such as convolutional neural networks (CNNs) from early on \cite{ICCV15_Khosla}, and advancements in the field have allowed researchers to predict image memorability with increasing precision \cite{needell2022embracing,squalli2018deep,leonardi2019image}. The inductive bias (the assumptions of the learning algorithms used to generalize to unseen data) of CNNs is inspired by knowledge about the primate visual system, and activations in the networks layers have, with some success, been used to explain neural activations \cite{yamins2014performance}. However, some vulnerabilities of CNNs have been noted. For example, CNNs appear to depend more on image texture than biological vision systems do \cite{baker2020local}, and have problems with recognizing images based on the shape of objects (e.g., when texture is suppressed or removed). However, this vulnerability is reduced when the model’s shape bias is increased through training on shape representations \cite{geirhos2018imagenet}.

The LaMem train/test splits is a well-established benchmark for memorability prediction \cite{ICCV15_Khosla}. The original MemNet \cite{ICCV15_Khosla}, which is based on AlexNet \cite{krizhevsky2012imagenet}, achieved a Spearman rank correlation of 0.64 on this benchmark. There have been several improvements on this benchmark, the leading approaches utilize image captioning to enhance memorability predictions. That is, a CNN produces a textual description of the image, which is then used to provide more high-level semantic information which is embedded into a semantic vector space before being combined with CNN image features in a multi-layered perceptron network. Squalli-Houssaini et al. \cite{squalli2018deep} used this approach to reach a Spearman correlation of 0.72, with a mean squared error (MSE) of approximately 0.0092 \cite{leonardi2019image}. Leonardi et al. \cite{leonardi2019image} used the captioning approach with dual ResNet50s and a soft attention mechanism to reach a rank correlation of 0.687 with an MSE of 0.0079. The ResMem model \cite{needell2022embracing}, which is a CNN-based residual neural network architecture (ResNet), uses LaMem, but also takes advantage of a more recently published dataset named MemCat \cite{goetschalckx2019memcat}. This is a data set containing 10,000 images based on categories of animals, food, landscape, sports and vehicles. This data set also has a higher split half correlation than LaMem. Needell and Bainbridge \cite{needell2022embracing} argue that the LaMem dataset on its own is lacking in generalizability due to poor sampling of naturalistic images. That is, the images are more intended as artistic renderings designed to attract an online audience. Hence by combining MemCat with LaMem this should potentially yield a more generalizable model. Moreover, the increased size of the combined dataset might help in driving the model performance further than previous models based on LaMem. The authors of ResMem also noted the importance of semantic information and structured their approach to utilize semantic representations from a ResNet model in order to improve predictions. An added benefit of ResMem is that it is shared on the python package index, which makes it easily accessible to researchers in diverse fields.

\subsection*{Vision transformers} Vision transformers (ViT) have recently been shown to provide similar or better performance than CNNs in a variety of computer vision tasks \cite{dosovitskiy2020image}. This architecture was first introduced in the natural language processing field \cite{vaswani2017attention} for capturing long-range dependencies in text. This architecture leads to superior speed/performance balance relativ to ResNet architectures \cite{he2016identity}. Moreover, ViTs have been shown to produce errors that are more similar to human errors \cite{tuli2021convolutional}, suggesting that they could take similar information into account (see also \cite{baker2022deep}). A reason for this may be that ViTs are likely to take more of the global context into account and be more dependent on the shape of objects rather than their texture \cite{tuli2021convolutional}. While it is not entirely clear why such properties may yield better predictions of image memorability, it could still help inform the discourse on the visual characteristics that are relevant as well as potentially yielding a better model for predicting image memorability. 

Hence, we set out to investigate if vision transformers can yield better predictions of memorability than the state-of-the-art in image memorability prediction. In particular, we aimed to (i) benchmark a model based on ViT against the well-established LaMem train/test splits \cite{ICCV15_Khosla}, (ii) train a ViT against the combined LaMem and MemCat data sets \cite{needell2022embracing} to benchmark against the ResMem model \cite{needell2022embracing}, (iii) train a final ViT model against a more diverse and deduplicated data set, (iv) validate the final ViT model against additional independent data sets and (v) inspect semantic level distributions of memorability scores for behavioral and predicted data. 

\section*{Methods}
As our model is based on ViT to predict memorability we named it ViTMem. Because it has been shown that low-level visual features are less important for image memorability prediction, it would seem appropriate to use image augmentations in training our ViTMem model to reduce overfitting. This approach have also been used by others \cite{leonardi2019image}, although not to the extent done here. The augmentations used consisted of horizontal flipping, sharpen, blur, motion blur, random contrast, hue saturation value, CLAHE, shift scale rotate, perspective, optical distortion and grid distortion \cite{buslaev2020albumentations}. For training all models we used PyTorch, the ADAM optimizer and mean squared error (squared L2 norm) for the loss function. Images were input as batches of 32 in RGB and resized to 256x256 pixels before applying augmentations with a probability of 0.7 and center cropping to 224x224 pixels. For creating ViTMem we used transfer learning on a vision transformer \cite{dosovitskiy2020image} model pretrained on ImageNet 1k (vit{\_}base{\_}patch16{\_}224{\_}miil) \cite{ridnik2021imagenet}. The final classification layer was reduced to a single output and a sigmoid activation function.

As we aim to provide an accessible model to the research community, it is also necessary to compare against the publicly available ResMem model. Unfortunately, the authors of ResMem did not publish their held-out test set, hence it is difficult to make a balanced comparison between the currently published ResMem model and any competing models. We propose to do 10 train/test splits that can be used by future researchers (available at https://github.com/brainpriority/vitmem{\_}data). Moreover, ResMem was not benchmarked on LaMem, hence a fair comparison can only be made on the combined LaMem and MemCat data set.

For the semantic level analysis, we chose to use image captioning \cite{wang2022unifying} as this provides an efficient method for deriving semantic properties from images at scale. Importantly, as the image captioning model was trained on human image descriptions, it is likely to extract content that humans find meaningful in images, and in particular objects and contexts that are relevant for conveying such meanings. Hence, nouns derived from such descriptions are likely to be representative portions of the content that would convey meaning to humans observing the images. 

\subsection*{Data Sources}
For the large-scale image memorability (LaMem) benchmark we used the LaMem dataset \cite{ICCV15_Khosla}. The image set used by ResMem is a combination of the image sets LaMem \cite{ICCV15_Khosla} and MemCat \cite{goetschalckx2019memcat}. LaMem containing 58,741 and MemCat 10,000 images, for a total of 68,741 images. ResMem is reported to have used a held-out test set with 5000 images, hence we randomly selected 5000 images as our test set for our 10 train/test splits for this combined data set. For our final model we aimed to clean up the data and combine more of the available data sets on image memorability. As number of duplicated images within and between data sets is unknown and duplicated images may interfere with performance measures, we aimed to deduplicate the data for this model. Duplicated images were identified by simply deriving embeddings from an off-the-shelf CNN model, and then visually inspecting the most similar embeddings. Our analysis of the data sets LaMem and MemCat showed that LaMem have 229 duplicated images while MemCat have 4. Moreover, 295 of the images in LaMem is also in MemCat.  We aimed to build a larger and more diverse data set by combining more sources, and for this we chose CVPR2011 \cite{isola2011understanding} and FIGRIM \cite{bylinskii2015intrinsic}. CVPR2011 had 6 internal duplicates, 651 duplicates against LaMem, 78 against MemCat og 9 against FIGRIM. FIGRIM had 20 duplicates against MemCat and 70 against LaMem. All identified duplicates were removed before merging the data sets. As the images from FIGRIM and CVPR2011 were cropped, we obtained the original images before including them in the data set. This resulted in a data set with 71,658 images. For this data set we performed a 10{\%} split for the test set.

\section*{Results}
\subsection*{Results on LaMem data set}

On the LaMem data set the ViTMem model reached an average Spearman rank correlation of 0.711 and an MSE of 0.0076 (see Table~\ref{tab:results1}). Here we compare our performance to measures obtained by MemNet \cite{ICCV15_Khosla}, Squalli-Houssaini et al. \cite{squalli2018deep} and Leonardi et al. \cite{leonardi2019image}.
\newline

\begin{table}[h]
 \caption{Comparison of model performance on LaMem data set}
  \centering
  \begin{tabular}{lcc}
    \toprule
    \cmidrule(r){1-3}
    Model & MSE Loss $\downarrow$ & Spearman $\rho$
    $\uparrow$ \\
    \midrule
    MemNet          & Unknown & 0.640  \\
    Squalli-Houssaini et al.  & 0.0092  & 0.720 \\
    Leonardi et al. & 0.0079  & 0.687 \\
    ViTMem          & 0.0076  & 0.711 \\
    \bottomrule
  \end{tabular}
  \label{tab:results1}
\end{table}

\subsection*{Results on the combined LaMem and MemCat data set}
Training on 10 train/test splits on the combined data set the results showed that ViTMem performed better than the ResMem model (see Table~\ref{tab:results2}). The average across splits showed a Spearman rank correlation of 0.77 and an MSE of 0.005.
\newline

\begin{table}[!h]
 \caption{Model performance on LaMem and MemCat combiend dataset}
  \centering
  \begin{tabular}{lcc}
    \toprule
    \cmidrule(r){1-3}
    Model & MSE Loss $\downarrow$ & Spearman $\rho$
    $\uparrow$ \\
    \midrule
    ResMem  & 0.009  & 0.67     \\
    ViTMem  & 0.005 & 0.77      \\
    \bottomrule
  \end{tabular}
  \label{tab:results2}
\end{table}

\subsection*{Results on combined and cleaned data set}

To assess model performance on the larger and cleaned data set, we made a train/test split and then performed repeated k-fold cross validation with 10 train/test splits on the training set. This resulted in a mean MSE loss of 0.006 and a mean Spearman rank correlation of 0.76 (see Table 3). In order to provide a model for the community we used the full data set to train the final model (ViTMem Final Model), which is published on the python package index as version 1.0.0. This was trained on the full training set and tested on its corresponding test set. The results showed a Spearman rank correlation of 0.77 and an MSE of 0.006 (see Table~\ref{tab:results3}). The train/test splits are available on github.
\newline

\begin{table}[!h]
 \caption{Model performance on combined and cleaned data set}
  \centering
  \begin{tabular}{lcc}
    \toprule
    \cmidrule(r){1-3}
    Model & MSE Loss $\downarrow$ & Spearman $\rho$
    $\uparrow$ \\
    \midrule
    ViTMem  & 0.006 & 0.76      \\
    ViTMem Final Model & 0.006 & 0.77      \\
    \bottomrule
  \end{tabular}
  \label{tab:results3}
\end{table}

\subsection*{Validation on independent data sets}
To further validate our model, we used memorability scores from an independent data set by Dubey and colleagues named PASCAL-S \cite{dubey2015makes, li2014secrets} consisting of 852 images and cropped objects from the same images. ViTMem achieved a Spearman correlation of 0.44 on the images and 0.21 on the objects. In comparison ResMem achieved a correlation of 0.36 on the images and 0.14 on the objects. Validating against the THINGS data set \cite{Kramer2022.04.29.490104}, which consists of 26,106 images with memorability scores, achieved a Spearman rank correlation of 0.30 for ViTMem and 0.22 for ResMem.

\subsection*{Semantic level analysis}
\begin{figure*}[tbhp]
    \centering
    \includegraphics[width=1.0\linewidth]{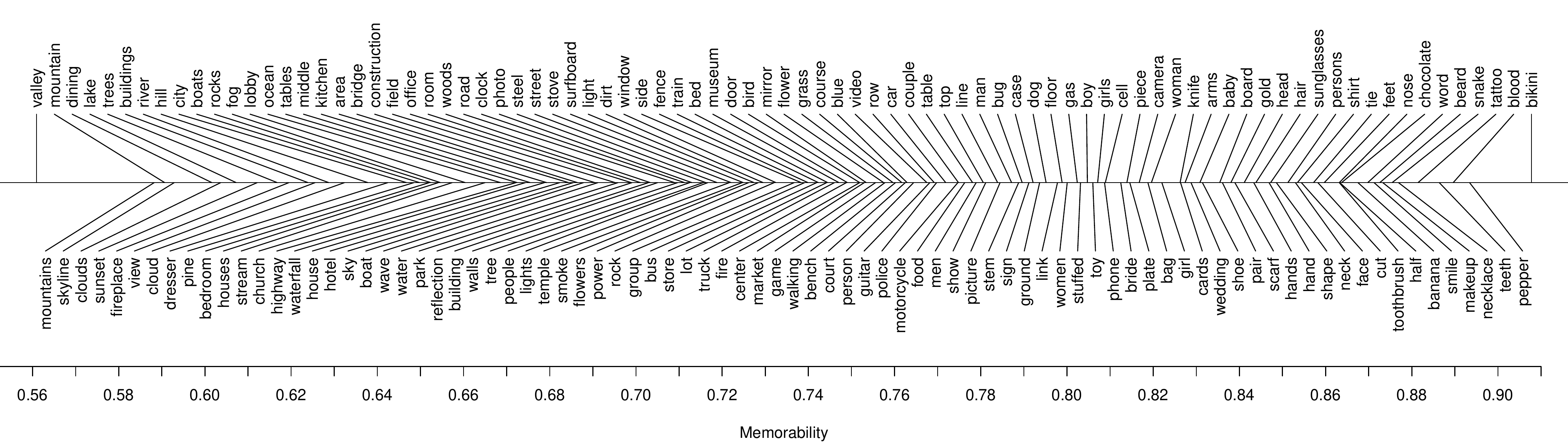}
    \caption{Average behavioral image memorability scores for nouns that were extracted from images in the LaMem and MemCat data sets. The nouns shown are those that occurred most frequently or that are more frequent in the English language \cite{robyn_speer_2022_7199437}.}
    \label{fig:nouns_beh}
\end{figure*}

\begin{figure*}[tbhp]
    \centering
    \includegraphics[width=1.0\linewidth]{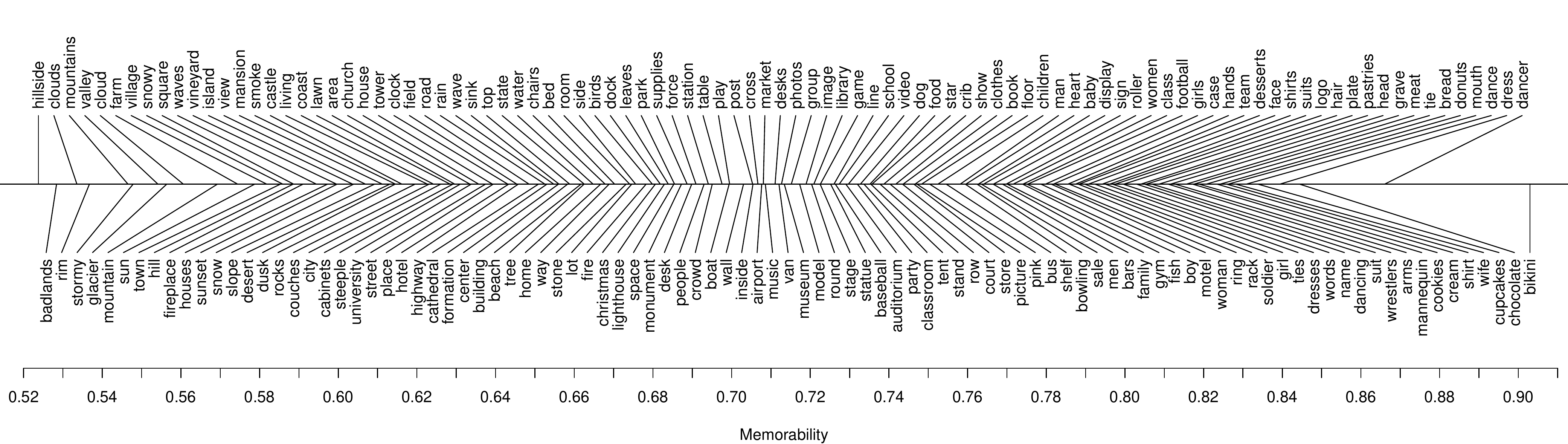}
    \caption{Average ViTMem predicted image memorability scores for nouns that were extracted from images in the Places205 data set. The nouns shown are those that occurred most frequently or that are more frequent in the English language \cite{robyn_speer_2022_7199437}.}
    \label{fig:nouns_pred}
\end{figure*}

\begin{figure}[tbhp]
    \centering
    \includegraphics[width=1.0\linewidth]{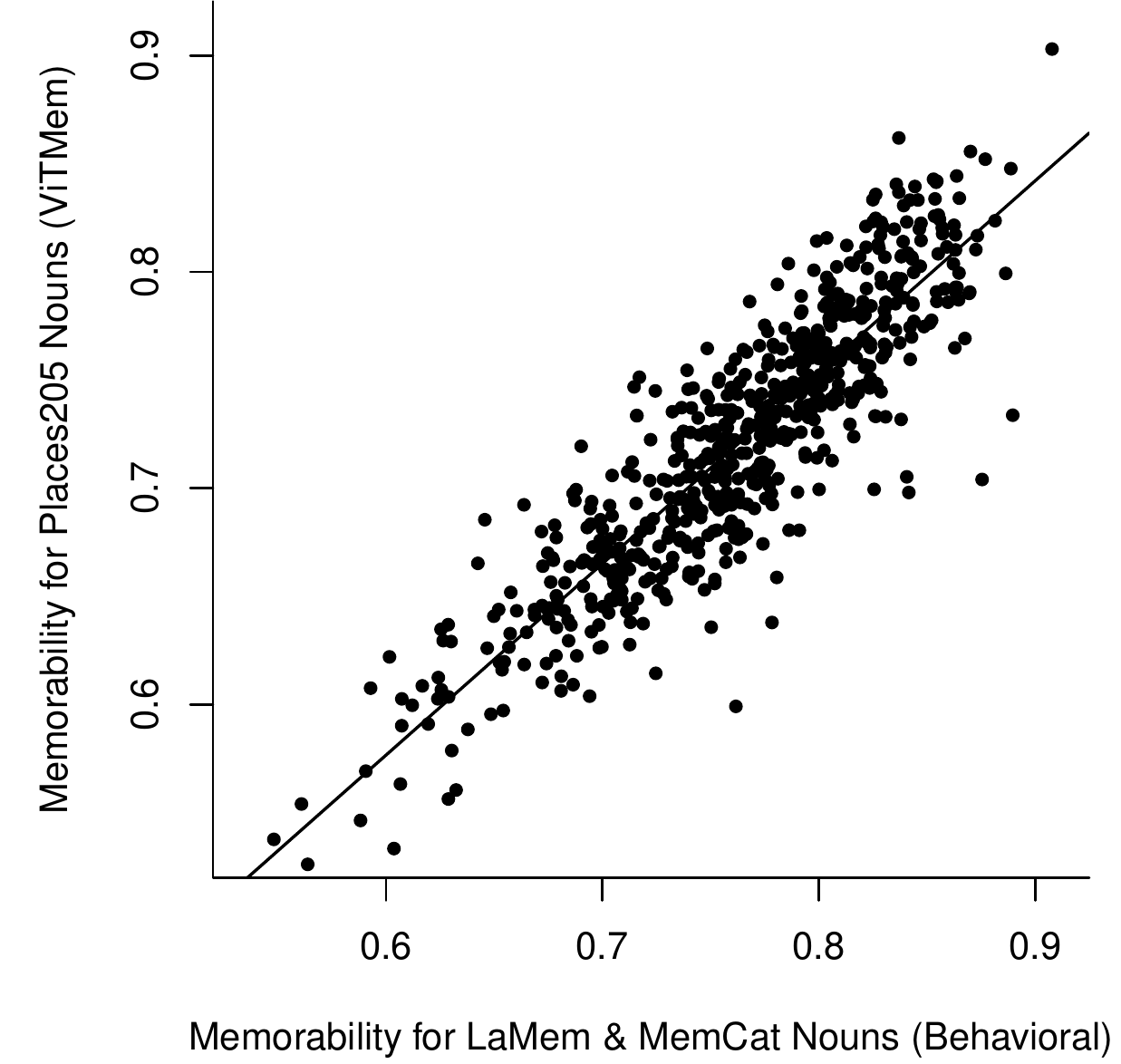}
    \caption{Average memorability scores for images with matching nouns in different data sets. The y-axis shows average predicted memorability scores from ViTMem on the Places205 data set. The x-axis shows average behavioral memorability scores on the combined LaMem and MemCat data set.}
    \label{fig:beh_pred}
\end{figure}

In order to better understand how the model predictions relate to the semantic content of the images, we performed image captioning \cite{wang2022unifying} on the combined LaMem and MemCat data set and the Places205 data set \cite{zhou2014learning}. We extracted nouns from the resulting image descriptions and averaged behavioral or predicted memorability scores for each noun \cite{loria2021textblob}. That is, the memorability for each image was assigned to each noun derived from the image captioning procedure. For the combined LaMem and MemCat data set we averaged behavioral memorability scores over nouns (see Figure \ref{fig:nouns_beh}), while for the Places205 data set we averaged predicted memorability scores from the ViTMem model (see Figure \ref{fig:nouns_pred}). A general interpretation of the visualizations in Figure 1 and 2 is that they appear to reveal a dimension from nouns usually observed outdoors to more indoor related nouns and ending with nouns related to animals, and in particular, humans. This would appear to reflect the distributions observed in previous work \cite{isola2011understanding, Kramer2022.04.29.490104}, and hence help to validate the model in terms of the image content it is sensitive to. To further investigate how well memorability associated with nouns were similar across the models we selected nouns occurring more than the 85th percentile in each set (654 nouns for LaMem and MemCat, 2179 nouns for Places205), this resulted in 633 matched nouns across sets. Analysis of these showed a Spearman ranked correlation of 0.89 and a R${\textsuperscript{2}}$ of 0.79, ${\emph{p}}$<0.001 (see Figure \ref{fig:beh_pred}). This analysis indicates that nouns from image captioning is a strong predictor of image memorability and that the ViTMem model is able to generalize the importance of such aspects from the training set to a new set of images.

\section*{Discussion}
Using vision transformers we have improved on the state-of-the-art in image memorability prediction. Results showed that ViTMem performed equal to or better than state-of-the-art models on LaMem, and better than ResMem on the LaMem and MemCat hybrid data set. In addition, we assembled a new deduplicated hybrid data set and benchmarked the ViTMem model against this before training a final model. The model was further validated on additional data sets, and performed better than ResMem on these as well. Finally, we ran a semantic level analysis by using image captioning on the hybrid data set. We ranked the behavioral memorability scores on the images, labeled with nouns extracted from the captioning procedure. The results revealed that images labeled by nouns related to landscapes, cities, buildings and similar, were ranked lowest, whereas images labeled by nouns related to animate objects and food, were ranked highest. This finding is consistent with known category effects on memorability \cite{bylinskii2015intrinsic, dubey2015makes, goetschalckx2019memcat, ICCV15_Khosla, Kramer2022.04.29.490104} and suggests that the labels extracted from captioning procedure is strongly related to factors that drive memorability for those images. Subsequently, we predicted memorability scores on images from a novel data set (Places205), ran the image captioning procedure, and ranked the predicted memorability scores on the images, labeled with nouns extracted from the captioning procedure. Visual inspection of the results revealed that the ranks were similar across samples and methods. This impression was confirmed by a strong correlation between matching pairs of nouns and 79${\%}$ explained variance, suggesting that ViTMem captures the semantic content that drives memorability in images.

The use of image augmentations in training the ViTMem model in combination with state-of-the-art performance suggest that such augmentations are not disrupting the ability of the model to predict image memorability and hence may further support the importance of semantic level properties in image memorability. That is, the augmentations modify a range of low-level image properties but mostly leave the semantic content intact.

In comparison with ResMem, which relies on a CNN-based residual neural network architecture, ViTMem is based on vision transformers which integrate information in a more global manner \cite{tuli2021convolutional}. As images are compositions of several semantically identifiable objects or parts of objects, a more holistic approach may be more apt at delineating the relative relevance of objects given their context. That is, we speculate that a broader integration of image features allows for a more complete evaluation of its constituent features in relation to each other. Hence, if semantic content is important for predicting image memorability, the model may have weighed the importance of semantic components in relation to each other to a larger degree than models based on CNNs.

ViTMem code and train/test sets are shared on github (https://github.com/brainpriority/), and a python package named vitmem is available on the python package index (see supplementary \cref{note:Note1} for a tutorial). Researchers and interested parties can use the model to predict memorability in existing or novel stimuli and employ them in research or applied settings. The ViTMem model will allow researchers to more precisely predict image memorability. The release of ViTMem follows up ResMem in providing an accessible method for predicting image memorability. This is important for studies aiming to control for how easily an image can be remembered. This will for example allow experimental psychologists and neuroscientists to better control their research. Similarly, educators, advertisers and visual designers can leverage the model to improve the memorability of their content.

Despite state-of-the-art performance in memorability prediction, improvements may still be possible to achieve. Previous works have shown benefits of pretraining their networks on data sets of places and objects prior to fine tuning for memorability prediction \cite{perera2019image}. Moreover, ViTMem do not take image captioning into account, which have been successfully done with CNNs \cite{squalli2018deep, leonardi2019image}. Hence there is potentially more to be gained from incorporating image semantics and/or pretraining on data sets of objects and places. In addition, ViTMem is only based on the "base" configuration of the available ViT models. Model performance may still increase by adopting the “large” or “huge” configurations of the model.

We conclude that ViTMem can be used to predict memorability for images at a level that is equal to or better than state-of-the-art models, and we propose that vision transformers provide a new step forward in the computational prediction of image memorability.

\section*{References}
\bibliography{Article.bib}

\onecolumn
\newpage

\captionsetup*{format=largeformat}
\section{How to use the vitmem python package} \label{note:Note1} 

Python needs to be installed on a computer before pip can be used to install the vitmem package.

To install vitmem from a command prompt run:

\begin{verbatim}
pip install vitmem
\end{verbatim}

To predict image memorability for an image named "image.jpg", run the following in a python interpreter:

\begin{verbatim}
from vitmem import ViTMem
model = ViTMem()
memorability = model("image.jpg")
print(f"Predicted memorability: {memorability}")
\end{verbatim}

\end{document}